\documentclass[twocolumn,letterpaper]{ieeeconf}

\usepackage{amsmath}
\usepackage{amssymb}
\usepackage{graphicx}
\usepackage{subfigure}
\usepackage{verbatim}
\usepackage[thinlines,thiklines]{easybmat}
\usepackage{latexsym}
\usepackage{cite}
\usepackage{stmaryrd}
\usepackage{multirow}
\usepackage{textcomp}
\usepackage[ruled]{algorithm2e}
\usepackage{diagbox}
\usepackage{array}
\usepackage{color,soul}
\usepackage{layout}
\sethlcolor{yellow} 

\newcolumntype{M}[1]{>{\centering\arraybackslash}m{#1}}
\newcolumntype{N}{@{}m{0pt}@{}}

\usepackage[left=0.77in,top=0.75in,right=0.77in,bottom=0.78in]{geometry}
\newbox\tempbox

\newcommand{\bs}[1]{\ensuremath{{\boldsymbol{#1}}}}

\newtheorem{rem}{Remark}
\begin{document}

\title{\bf \Large Contingency Model Predictive Control for Bipedal Locomotion on Moving \\
	Surfaces with a Linear Inverted Pendulum Model~\thanks{The work was supported in part by the US NSF under award CMMI-2222880 (Yi).}}

\author{Kuo Chen\thanks{K. Chen is with Motional AD, LLC, 100 Northern Ave, Boston, MA 02210 USA (email:kc625@scarletmail.rutgers.edu).}, Xinyan Huang, Xunjie Chen, and Jingang Yi\thanks{X. Huang, X. Chen, and J. Yi are with the Department of Mechanical and Aerospace Engineering, Rutgers University, Piscataway, NJ 08854 USA (email: xh301@rutgers.edu, xc337@rutgers.edu, jgyi@rutgers.edu).}}

\maketitle
\pagestyle{empty}  
\thispagestyle{empty} 

\begin{abstract}
Gait control of legged robotic walkers on dynamically moving surfaces (e.g., ships and vehicles) is challenging	due to the limited balance control actuation and unknown surface motion. We present a contingent model predictive control (CMPC) for bipedal walker locomotion on moving surfaces with a linear inverted pendulum (LIP) model. The CMPC is a robust design that is built on regular model predictive control (MPC) to incorporate the ``worst case'' predictive motion of the moving surface. Integrated with an LIP model and walking stability constraints, the CMPC framework generates a set of consistent control inputs considering to anticipated uncertainties of the surface motions. Simulation results and comparison with the regular MPC for bipedal walking are conducted and presented. The results confirm the feasibility and superior performance of the proposed CMPC design over the regular MPC under various motion profiles of moving surfaces.
\end{abstract}

\section{Introduction}
\label{Sec:intro}

Humanoid robots inevitably walk on various terrains, including moving surfaces such as ships and vehicles. Gait control of legged robotic walkers on these dynamically moving surfaces is challenging due to the limited balance control actuation of intrinsically unstable bipedal system and unknown surface motion. Although many modeling and control designs have been reported for bipedal walking on firm stationary surface, few are developed and reported for dynamically moving surfaces. Bipedal walkers are commonly modeled as multi-link rigid body and hybrid zero dynamics were usually used to design low-level walking gait controllers~\cite{westervelt2018feedback}. For high-level gait planning and foot placement design, reduced-order robotic walking models, such as linear inverted pendulum (LIP) model~\cite{kajita2003biped}, have been used extensively because of its simplicity to reveal the motion relationship between the center of mass (COM) and the zero moment point (ZMP) of the bipedal robots. Variations of LIP models are explored to different scenarios by considering additional degree-of-freedom (DOF), such as foot slip~\cite{chen2015robotic,mihalec2020recoverability,MitjaSlipDetection2019, trkov2019bipedal}, two-mass LIP model~\cite{mihalec2021recoverability,MihalecTMech2023}, and hybrid dynamics~\cite{xiong20223D}.

For stable walking gait design, divergent component of motion (DCM)~\cite{englsberger2015three} and capture point (CP)~\cite{pratt2006capture} were proposed to highlight the unstable mode of the LIP model for stable ZMP generation. The work in~\cite{lanari2014boundedness} proposed an initial condition of ZMP control that ensured the converged and stable solution of the capture point. In~\cite{scianca2016intrinsically}, an intrinsically stable model predictive control (MPC) was designed with the stability constraints that were developed in~\cite{lanari2014boundedness}. MPC is able to handle input and state constraints and has been used to design bipedal robot control under uncertainties~\cite{villa2017model,wieber2006trajectory}. Robust MPC design were also reported (e.g.,~\cite{PandalaRAL2022,YeganegiTRO2022}) for quadrupedal and bipedal robots under unmodelled dynamics or disturbances. The research approach in~\cite{xu2023Robust} used a robust convex MPC framework to handle uncertain friction constraints and uncertain model dynamics for quadruped walking. A linear stochastic MPC was discussed in~\cite{gazar2021stochastic} accounting for uncertainties from the bipedal walking dynamics and comparative studies were conducted to evaluate its performance with the tube-based robust MPC. However, all above studies were built on the assumption of rigid, stationary surfaces.

Passivity-based whole-body dynamics control was developed for humanoid robots on limited movable surfaces such as deformable surfaces~\cite{HenzeRAL2018}. Reinforcement learning approach was recently reported for humanoid robot control on rotating platform~\cite{XiACCESS2020}. The recent work in~\cite{gao2023time} aimed at developing control strategy for bipedal robot walking on a swaying rigid surface with known periodic motion. Iqbal~{\em et~al.}~\cite{iqbal2020provably} developed a walking controller for a quadrupedal robot that explicitly addressed the hybrid, nonlinear and time-varying robot dynamics. An online foot placement strategy was proposed in~\cite{iqbal2021extended} and the design was based on the extended CP concept for moving surfaces. The above studies addressed control strategies using the surface motion estimated at the current time without explicitly incorporating predictive information.

Predictive methods are helpful for capturing probability of uncertainties and many above-mentioned stochastic MPC methods suffer over-conservatism for control design. In~\cite{zhan2016non}, a probabilistic behavior modeling of the non-conservatively defensive strategy was proposed to avoid overcautious actions of vehicle agents at the road intersection. The autonomous driving strategy developed in~\cite{alsterda2019contingency} further used a new contingency model predictive control (CMPC) approach that can be selectively robust to emergencies at the road interaction. One advantage of the contingency concept was to enable the controller anticipate events that might take place instead of reacting when emergencies occurred, which achieved responsible but practical conservatism.

Inspired by the above discussed autonomous driving strategy at road interaction, we present a CMPC design for bipedal walkers on moving surfaces. We consider an LIP model for bipedal walkers to illustrate and demonstrate the CMPC method. An alternative LIP model is first presented to emphasize the divergent component as a stability constraint of the CMPC. Instead of assuming a known periodic time-varying motion (e.g.,~\cite{gao2023time,iqbal2021extended}), the surface movement is assumed to follow bounded acceleration and bounded jerk. We conduct simulation studies to demonstrate the new CMPC design and also compare with the regular MPC in literature. The main contribution of this work lies in the novel CMPC approach to incorporate the predictive motion of the moving surface under ``worse'' case situation. To the best authors' knowledge, for the first time, this work adopts the contingency concept to estimate the envelopes of the surface motion (i.e., uncertainties) and integrates such estimation into MPC design to stabilize bipedal walking locomotion. Compared with the other robust MPC approaches, the main advantage of the CMPC approach lies in its less conservatism and computational efficacy with guaranteed performance.


\section{Bipedal LIP Model on Moving Surface}
\label{Sec:ProbConfig}

\begin{figure}[h!]
	\centering
	\includegraphics[width=3.3in]{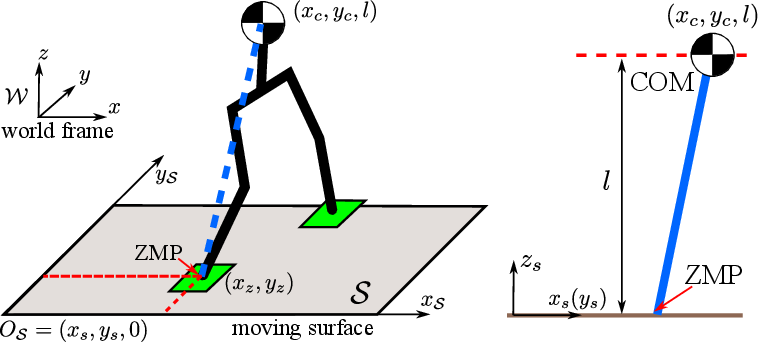}
	\caption{Schematic of bipedal robotic walker on a moving surface frame $\mathcal{S}$ (left) and an LIP model (right). }
	\label{fig:WalkingLIPM}
\end{figure}

Fig.~\ref{fig:WalkingLIPM} illustrates the schematic and configuration of a bipedal robot walking on a moving surface, denoted by $\mathcal{S}$. We consider a world coordinate frame $\mathcal{W}$ and a local frame is fixed on $\mathcal{S}$ with the original point $O_{\mathcal{S}}$ at $(x_s,y_s,0)$ in $\mathcal{W}$. To simplify the problem, the motion of surface $\mathcal{S}$ is only in horizontal ($x,y$) directions. We assume that the bipedal robot's foot never slips on $\mathcal{S}$, that is, the velocity of the robotic foot is the same as that of the moving surface $\mathcal{S}$ as long as the foot is in contact with the surface.

The bipedal walker is modeled as an LIP with constant COM height, denoted by $l$. The walking process consists of two phases: single-stance and double-stance phases. The motion of COM can be expressed through an inverse pendulum model pinned on ZMP. We present the LIP model in frame $\mathcal{S}$. Denoting the COM and ZMP positions as $(x_c,y_c)$ and $(x_z,y_z)$ in $\mathcal{S}$, respectively, as shown in Fig.~\ref{fig:WalkingLIPM}, the motion of COM is expressed as~\cite{gao2023time}
\vspace{-2mm}
\begin{equation}
\ddot{x}_c= \omega^2 (x_c -x_z) - \ddot{x}_s,\; \ddot{y}_c= \omega^2 (y_c - y_z) - \ddot{y}_s,
\label{eqn:2D_LIPM}
\end{equation}
where $\omega = \sqrt{g/l}$ and $g$ is the gravitational constant. In~\eqref{eqn:2D_LIPM}, the motions in two directions are decoupled and therefore, in the following derivation, we first consider the $x$-direction motion. The conclusion can be extended similarly to the $y$-direction.

We first define a coordinate transformation such that
\begin{equation}
\label{eqn:xi_var}
\xi_u = x_c + \frac{1}{\omega} \dot{x}_c,\quad \xi_s = x_c - \frac{1}{\omega} \dot{x}_c,
\end{equation}
where $\xi_u$ is the divergent (unstable) component while $\xi_s$ is the stable component~\cite{englsberger2015three,pratt2006capture}. By substituting~\eqref{eqn:xi_var} into~\eqref{eqn:2D_LIPM}, we obtain
\begin{equation}
\label{eqn:xi_dyn}
    \dot{\xi}_u = \omega (\xi_u - x_z - \frac{1}{\omega^2} \ddot{x}_s),~
    \dot{\xi}_s = -\omega (\xi_s - x_z - \frac{1}{\omega^2} \ddot{x}_s).
\end{equation}
By defining the effective ZMP as $x_e= x_z + \frac{1}{\omega^2}\ddot{x}_s$, \eqref{eqn:xi_dyn} is rewritten as
\begin{equation}\label{eqn:LIPM_xi}
    \dot{\xi}_u = \omega (\xi_u - x_e),\quad
    \dot{\xi}_s = -\omega (\xi_s - x_e).
\end{equation}
\eqref{eqn:LIPM_xi} is regarded as an alternative form of LIP model of the bipedal walking whose stable and unstable modes are decoupled.

The effective ZMP position $x_e$ is the input of the LIP model in~\eqref{eqn:LIPM_xi}. We treat the component $x_z$ as the control variable that should be optimized for stability. The motion of the moving surface $x_s(t)$ is treated as a source of unknown disturbance. We assume that the acceleration $\ddot{x}_s$ can be measured at the current time.

The solution of the unstable component $\xi_u(t)$ in~\eqref{eqn:LIPM_xi} is
$\xi_u(t) = \xi_u(t_0) e^{\omega (t-t_0)} - \omega \int_{t_0}^{t} e^{\omega(t-\tau)} x_e(\tau) d\tau$.
However, with a special choice of initial state
\begin{equation}\label{eqn:stability_condition1}
    \xi_u(t_0) = \omega \int_{t_0}^\infty e^{-\omega(\tau - t_0)}x_e(\tau)d\tau,
\end{equation}
the final value $\xi_u(t)=\omega \int_{t}^{\infty}e^{-\omega(\tau - t)}x_e(\tau)d\tau$ is stable~\cite{scianca2016intrinsically,lanari2014boundedness}. This choice of initial condition~\eqref{eqn:stability_condition1} is called the stability condition. It is a noncausal condition because the initial state at $t_0$ relates to the effective ZMP trajectory in the future. The treatment of the stable initial condition as one constraint for model predictive control will be discussed in the next section.

\section{Contingency MPC for Bipedal Walking}
\label{Sec:CMPC}

In this section, we first present the estimate of bounded ranges of unknown disturbance and the predictive estimation is then integrated with the CMPC framework.
\vspace{-3mm}
\subsection{Boundedness of Unknown $\ddot{x}_s(t)$}

The motion of the moving surface $x_s(t)$ is considered as an unknown disturbance of the LIP model~\eqref{eqn:LIPM_xi}. At current time step $t_k$, $k \in \mathbb{N}$, acceleration measurement $\ddot{x}_s(t_k)$ is available. For most applications, it is reasonable to assume that: (1) the acceleration $\ddot{x}_s(t)$ is bounded within the control horizon period, that is,  $\ddot{x}_s(t) \in [a_{\min}, a_{\max}]$ for $ t \in [t_k,t_k+T_c]$, where $T_c$ denotes the control predictive horizon, $a_{\min}$ and $a_{\max}$ are the lower and upper bounds of $\ddot{x}_s(t)$, respectively. (2) The jerk of the moving surface is also bounded, namely, $\dddot{x}_s(t)\in [j_{\min}, j_{\max}]$ for $t \in [t_k,t_k+T_c]$, where $j_{\min}$ and $j_{\max}$ are the minimum and maximum values of the jerk, respectively.

\begin{figure}[h!]
	\centering
	\includegraphics[width=2.7in]{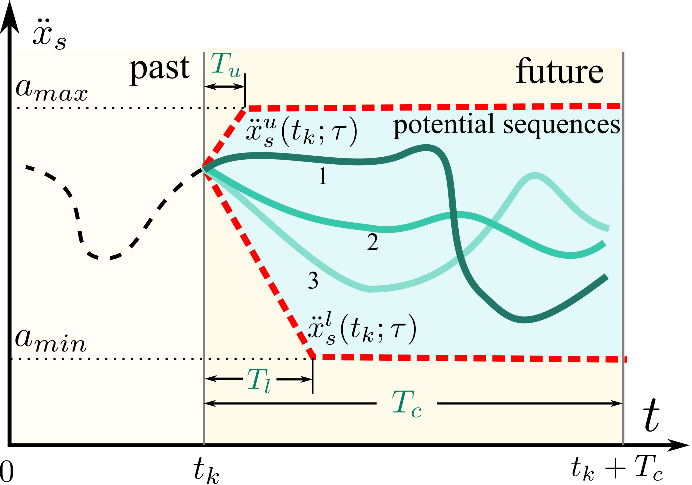}
	\caption{The schematic of acceleration $\ddot{x}_s$ calculation at $t_k$ within the MPC control horizon $T_c$.}
	\label{fig:bounded_disturbance}
	\vspace{-4mm}
\end{figure}

With the above assumptions, we can compute a bounded range of all possible trajectories of $\ddot{x}_s(t)$ for $t \in [t_k,t_k+T_c]$. For presentation convenience, we introduce the notation $x_s(t_k;t)=x_s(t_k+t)$ for $x_s$ and other variables. We then compute the low and upper bounds, denoted respectively by $\ddot{x}_{s}^l(t_k;\tau)$ and $\ddot{x}_{s}^u(t_k;\tau)$ of $\ddot{x}_{s}(t_k;\tau)$, $\tau \in [0,T_c]$, as follows.
\begin{align*}
  \ddot{x}_{s}^l(t_k;\tau) & =
  \begin{cases}
    \ddot{x}_s(t_k)+j_{\min}\tau, & \mbox{if } \tau < T_l \\
    a_{\min}, & \mbox{if } \tau \geq T_l.
  \end{cases} \\
  \ddot{x}_{s}^u(t_k;\tau) & =
  \begin{cases}
    \ddot{x}_s(t_k)+j_{\max}\tau, & \mbox{if } \tau < T_u \\
    a_{\max}, & \mbox{if } \tau \geq T_u,
  \end{cases}
\end{align*}
where $T_l$ and $T_u$ are the time constants for $\ddot{x}_{s}(t_k)$ to reach $a_{\min}$ and $a_{\max}$, respectively. It is calculated as $T_l = ~\frac{a_{\min}-\ddot{x}_s(t_k)}{j_{\min}},~T_u = \frac{a_{\max}-\ddot{x}_s(t_k)}{j_{\max}}$. Fig.~\ref{fig:bounded_disturbance} illustrates the above calculation and variable definitions. Therefore, the acceleration of the moving surface is constrained within the region as shown in the figure, that is, $\ddot{x}_{s}^l(t_k;\tau) \leq \ddot{x}_s(t_k;\tau) \leq \ddot{x}_{s}^u(t_k;\tau)$ for $\tau \in [0,T_c]$.

\subsection{CMPC Formulation}
\vspace{-0mm}

\subsubsection{Discretization of control vector}

To present MPC formulation in a compact form, discretization of the ZMP control input is considered within the control horizon $T_c$. At current time step $t_k$, we denote the discrete ZMP control sequence vector as $\dot{\bs{U}}_k\in \mathbb{R}^N$ as
\vspace{-0mm}
\begin{equation*}
  \dot{\bs{U}}_k = [\dot{x}_{z,k}\,\cdots \, \dot{x}_{z,k+i}\,\cdots\,\dot{x}_{z,k+N-1}]^T,
\end{equation*}
where $\dot{x}_{z,k+i}=\dot{x}_z(t_k;i\delta t)$ with the zero-order hold rule for time $t \in [t_k+i\delta t,t_k+(i+1)\delta t]$, $i=0,1, \cdots,N-1$, and $N \in \mathbb{N}$ is the number of the discretization time slices within $T_c$. It is straightforward to obtain that $\delta t = T_c/N$.

We continue to define two specific ZMP control vectors $\dot{\bs{U}}_{k}^u$ and $\dot{\bs{U}}_{k}^l$ corresponding to two ``worst'' cases of the disturbance (i.e., $\ddot{x}_{s}^u(t_k;\tau)$ and $\ddot{x}_{s}^l(t_k;\tau)$), that is,
\vspace{-0mm}
\begin{equation}
\label{eqn:twoWorstCases}
\dot{\bs{U}}_{k}^u=\dot{\bs{U}}_{k}\Big{|}_{\ddot{x}_{s}^u},~ \dot{\bs{U}}_{k}^l=\dot{\bs{U}}_{k}\Big{|}_{\ddot{x}_{s}^l}.
\end{equation}
We enforce the equality constraints for two extreme cases to eliminate the ambiguity for executing controlling, that is,
\begin{equation}
\label{eqn:CMPC_constraint}
  \bs{S} \dot{\bs{U}}_{k}^{u} = \bs{S} \dot{\bs{U}}_{k}^{l},
\end{equation}
where $\bs{S} \in \mathbb{R}^{n\times N}$ is the block matrix that is taken from the first $n$ rows of $N \times N$ identity matrix $\bs{I}_{N}$. The constraint in~\eqref{eqn:CMPC_constraint} enforces the first $n$ steps of $\dot{\bs{U}}_{k}^l$ and $\dot{\bs{U}}_{k}^u$ are exactly the same. The intuition of CMPC is that if we can find $\dot{\bs{U}}_{k}^u$ and $\dot{\bs{U}}_{k}^l$ that stabilize the robot under two extreme scenarios shown in Fig.~\ref{fig:bounded_disturbance} and the beginning parts of the two input trajectories are the same, then the shared part of the two trajectories is the desired control sequence.

\subsubsection{Stability constraints}

As discussed at the end of Section~\ref{Sec:ProbConfig}, the choice of the control input trajectory $x_e$ should ensure the convergence of $\xi_u(t)$ by satisfying the stability condition~\eqref{eqn:stability_condition1}. Therefore, the state at the current (initial) time $t_k$ should satisfy that
\begin{eqnarray}
  \xi_u(t_k) &=& \omega \int_{0}^\infty e^{-\omega \tau} x_e(t_k + \tau) d\tau \nonumber \\ 
  & \approx & \omega\int_0^{T_c}e^{-\omega \tau}\left[x_z(t_k+\tau)+ \frac{\ddot{x}_s(t_k+\tau)}{\omega^2}\right] d\tau \nonumber \\ 
  &=& \omega \sum_{i=0}^{N-1}x_{z,k+i}\int_{i\delta t}^{(i+1)\delta t}e^{-\omega \tau} d\tau+b_{k}^j, 
\label{eqn:Stability}
\end{eqnarray}
where for different disturbance $\ddot{x}_{s}^j(t)$, $j=l,u$, the last step is obtained by defining and using
\begin{equation}
	b_{k}^j=\omega \int_{0}^{T_c} \frac{e^{-\omega\tau}}{\omega^2} \ddot{x}_{s}^j(t_k;\tau) d\tau, \quad j=l,u.
	\label{coeff1}
\end{equation}
The first term in~\eqref{eqn:Stability} can be re-written in terms of $\dot{\bs{U}}_k$. Corresponding to two extreme cases, the above stability constraints should be satisfied and there, we formulate that
\begin{equation}
\label{eqn:Stability_constraint}
    \xi_u(t_k) = \bs{A}_k^T \bs{U}_{k}^u + b_{k}^u = \bs{A}_k^T \bs{U}_{k}^l+ b_{k}^l,
\end{equation}
where $\bs{A}_k = (1 - e^{-\omega \delta t}) [1\;\,e^{-\omega \delta t}\;\,\cdots\;\,e^{-\omega (N-1) \delta t}]^T$.

\subsubsection{Foot placement and ZMP geometry constraints}

\begin{figure}[h!]
\vspace{-1mm}
	\centering
	\includegraphics[width=2.7in]{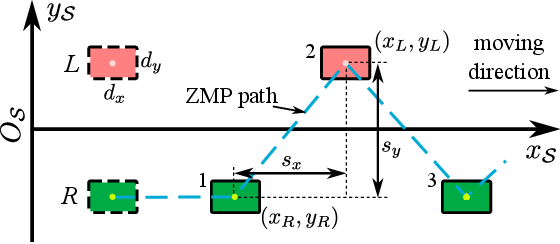}	
\vspace{-2mm}
	\caption{Schematic of step and ZMP configuration of bipedal walker.}
	\label{fig:StepConfig}
\vspace{-0mm}
\end{figure}

Fig.~\ref{fig:StepConfig} shows a symmetric foot placement of two legs. The robot foot is flat rectangular shape and the size is $d_x \times d_y$. The foot placement is determined by the center position of the foot, denoted by $(x_L,y_L)$ and $(x_R,y_R)$ for the left and right foot, respectively. The bipedal robot walks with the constant strike length ($s_x, s_y$) in both the $x$- and $y$-directions. To avoid making the robot falling, the ZMP should be placed in the stance foot polygon called ZMP admissible region, $\mathcal{P}=[x_{\min}(t),x_{\max}(t)]\times [y_{\min}(t),y_{\max}(t)]$, where $x_{\min}(t)$, $x_{\max}(t)$, $y_{\min}(t)$, and $y_{\max}(t)$ are minimum and maximum admissible boundaries in the $x$- and $y$-direction, respectively. Region $\mathcal{P}$ is the area about the foot placement and this implies that $\mathcal{P}$ should consider both the single- and double-stance phases.

For the CMPC regarding to the extreme cases, each ZMP control input within $T_c$ holds that   $x_{z,k+i}\big{|}_{\ddot{x}_{s}^u} \in \mathcal{P},~ x_{z,k+i}\big{|}_{\ddot{x}_{s}^l} \in \mathcal{P}$, that is,
\begin{equation}
\label{eqn:Geo_constraint}
\bs{U}_{k}^{\min} \leq \bs{U}_{k}^l \leq \bs{U}_{k}^{\max}, \;\, \bs{U}_{k}^{\min} \leq \bs{U}_{k}^u \leq \bs{U}_{k}^{\max},
\end{equation}
where $\bs{U}_{k}^{\min}$ and $\bs{U}_{k}^{\max}$ denote the upper and lower bounds of the control vectors and ``$\leq$'' in~\eqref{eqn:Geo_constraint} represents the element-wise comparison between two vectors. The ZMP input should satisfy the geometry constraint~\eqref{eqn:Geo_constraint} at each time slice.

\subsubsection{Optimization formulation}

We summarize the CMPC formulation for the $x$-direction motion. At $t_k$, we solve a receding horizon optimization problem with the decision variables $\bs{U}_{k}^l$ and $\bs{U}_{k}^u$ subjective to the stability and geometry constraints. We consider to minimize the ZMP change rate for two extreme cases such that a designed objective function is $J=\sum_{i=1}^{N-1}\left(x_{z,k+i}-x_{z,k+i-1}\right)^2$. Therefore, the optimization problem is formulated as
\begin{subequations}
\label{eqn:CMPC}
\begin{align}
\text{\hspace{-3mm}}    \min_{\bs{U}_{k}^u, \bs{U}_{k}^l} & J = \| \bs{C}\bs{U}_{k}^u \|^2 + \| \bs{C}\bs{U}_{k}^l \|^2, \\
    \text{s.t.}~& \bs{S} \bs{U}_{k}^u = \bs{S} \bs{U}_{k}^l,\\
    & \xi_{u}(t_k) = \bs{A}_k^T\bs{U}_{k}^u+b_{k}^u = \bs{A}_k^T\bs{U}_{k}^l+b_{k}^l,\\
    &\bs{U}_{k}^{\min} \leq \bs{U}_{k}^u \leq \bs{U}_{k}^{\max}, \;
    \bs{U}_{k}^{\min} \leq \bs{U}_{k}^l \leq \bs{U}_{k}^{\max},
\end{align}
\end{subequations}
where differentiation matrix $\bs{C} \in \mathbb{R}^{(N-1) \times N} $ is given as
\begin{equation*}
    \bs{C} =
\begin{bmatrix}
  1&  -1&  0&  ...& 0\\
  0&  1&  -1&  ...&0 \\
  \vdots&  \vdots&  \vdots&  \vdots& \vdots\\
  0&  0&  ...&  1&-1
\end {bmatrix}.
\end{equation*}
\rem Since motions in $x$ and $y$ direction are decoupled, the contingency MPC problem discussed in $x$ direction can be extended directly to $y$ direction. We take the first $n$ steps of the control prediction vectors as the actual ZMP control input and in this work, $n = 1$ was implemented.

\vspace{-0mm}
\section{Simulation Results}
\label{Sec:Sim}

We conducted simulation studies to validate and demonstrate the proposed CMPC design. A NAO robot was considered with the same LIP model parameters  as described in~\cite{scianca2016intrinsically}. Table~\ref{tab:parameters} lists the main model parameters and their values. The robot began to walk at the double stance and stepped the right foot first and then the left foot periodically with the same strike length. The foot placement location and timing was pre-determined. A straight walking was considered and the strike length were set as $s_x=5$~cm and $s_y=10$~cm in the $x$ and $y$ direction, respectively. The size of the foot was $d_x \times d_y= 2 \times 2$~cm. A full walking cycle consisted of one step of the left foot swing, one step of the right foot swing, and two double stances. The portion ratio between the single- and double-stance phases was $2:1$, that is, single-stance $66.7$~\% and double-stance phase $33.3$~\%, and the total walking cycle period was $T=0.3$ s. The total simulation included seven steps.

\renewcommand{\arraystretch}{1.35}
\setlength{\tabcolsep}{0.06in}
\begin{table}[h!]
\vspace{2mm}
\centering
\caption{The LIP model parameters and their values (from~\cite{scianca2016intrinsically}).}
\label{tab:parameters}
\begin{tabular}{ccccccc}
\hline\hline
$l$ (cm)         & $\omega$ (rad/s) & $s_x$ (cm)  & $s_y$  (cm) & $d_x(d_y)$ (cm)  & $T_c$ (s)  & $N$ \\ \hline
$26$ & $6.14$   & $5$ & $10$ & $2$ & $1$ & 100 \\ \hline\hline
\end{tabular}%
\vspace{-1mm}
\end{table}

Two types of MPC controllers were implemented and compared for robot walking performance. The CMPC controller (denoted as $\mathcal{C}_1$)  and an intrinsically stable MPC controller in~\cite{scianca2016intrinsically} (denoted as $\mathcal{C}_2$) were used for the ZMP generation. We modified $\mathcal{C}_2$ such that the current acceleration was measured and in the prediction horizon it was assumed that the acceleration kept unchanged. Both MPC controllers were implemented in Matlab and solved by \verb"quadprog". The control parameters in implementation included: $T_c=1$~s, $j_{\min}=-1$~m/s$^3$, $j_{\max}=1$~m/s$^3$ for the $x$ direction, while $j_{\min}=-2$~m/s$^3$, $j_{\max}=2$~m/s$^3$ for the $y$ direction, $N=100$, and $\delta t = T_c/N =0.01$~s. $\bs{U}_k^{\min}$ and $\bs{U}_k^{\max}$ were determined by the foot placement geometry constraints.

\begin{figure}[h!]
\vspace{-2mm}
	\centering
	\subfigure[]{
		\label{fig:sineDisturbance}
		\includegraphics[width=2.8in,height=1.85in]{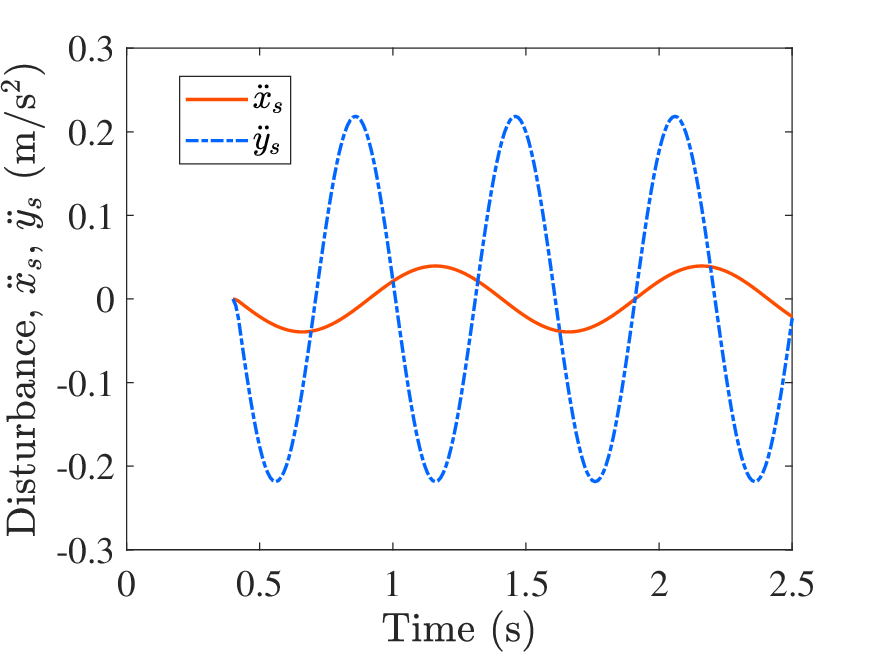}}
	\subfigure[]{
		\label{fig:randomDisturbance} 
		\includegraphics[width=2.8in,height=1.85in]{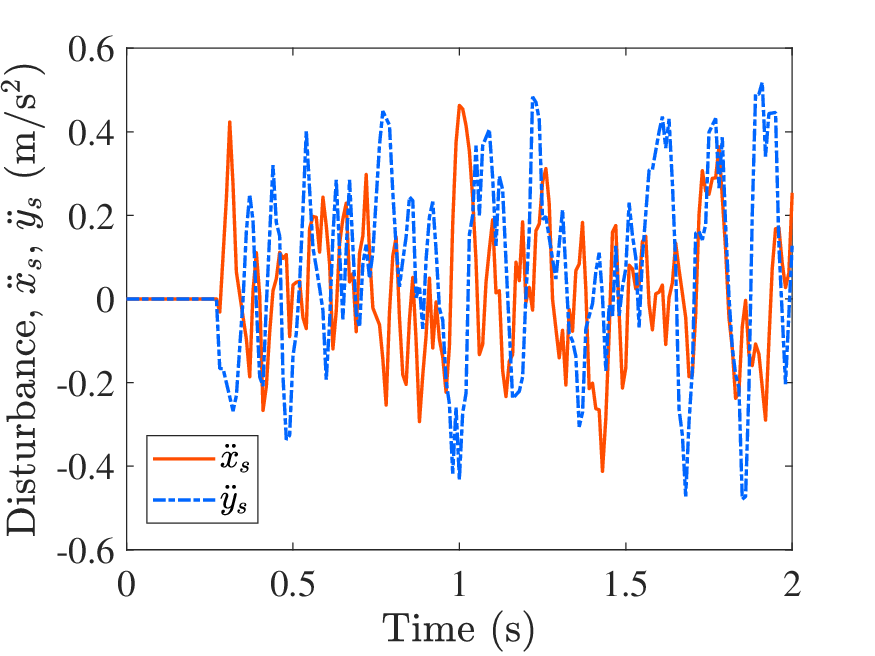}}
	\vspace{-1mm}	
	\caption{Two types of surface motion disturbances (red: $x$-direction and blue: in $y$-direction). (a) Sinusoidal disturbance. (b) Random disturbance.}
	\label{fig:disturbances}
	\vspace{-0mm}
\end{figure}

\begin{figure*}[t!]
\vspace{0mm}
\hspace{-1mm}
	\subfigure[]{
		\label{fig:ZMP:xResultSinDisturbance} 
		\includegraphics[width=2.42in]{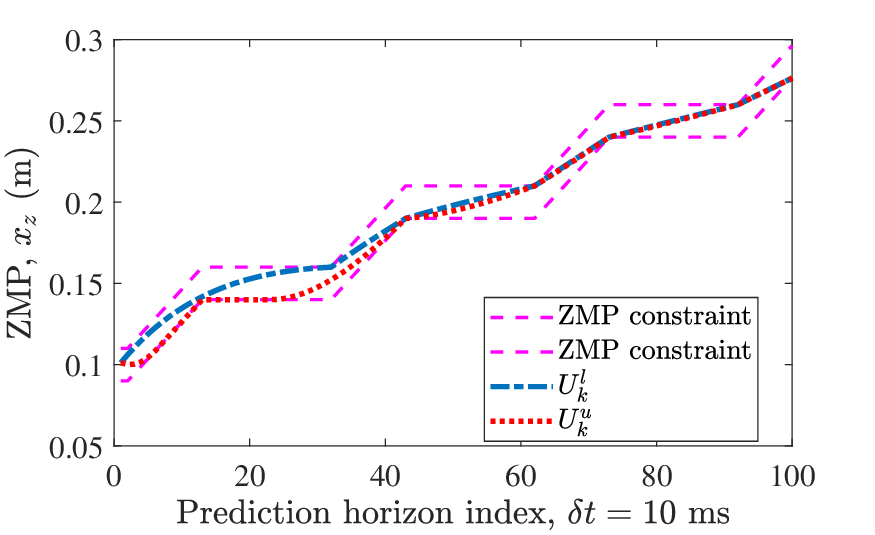}}
	\hspace{-7mm}
	\subfigure[]{
		\label{fig:ZMP:yResultSinDisturbance}
		\includegraphics[width=2.42in]{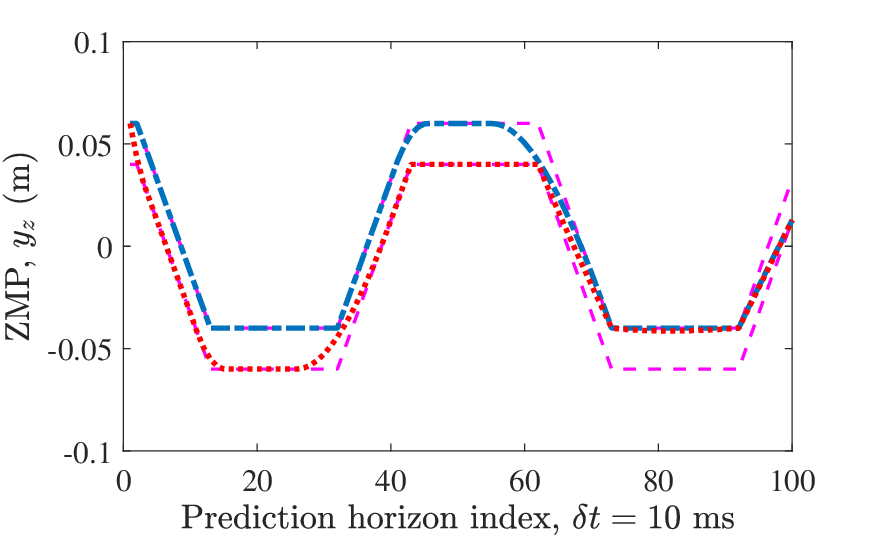}}
	\hspace{-7mm}
	\subfigure[]{
		\label{fig:ZMP:xyResultSinDisturbance}
		\includegraphics[width=2.43in]{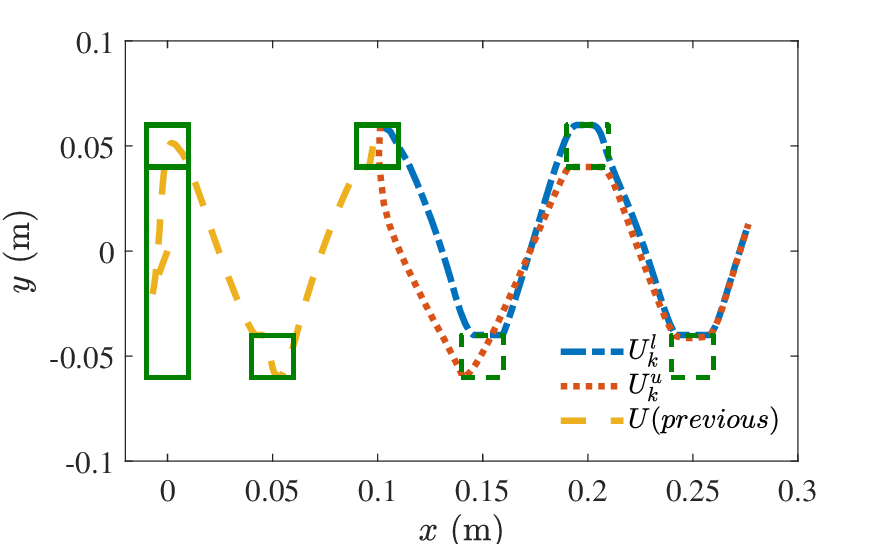}}
	\subfigure[]{
		\label{fig:ZMP:xResultRandomDisturbance} \includegraphics[width=2.42in]{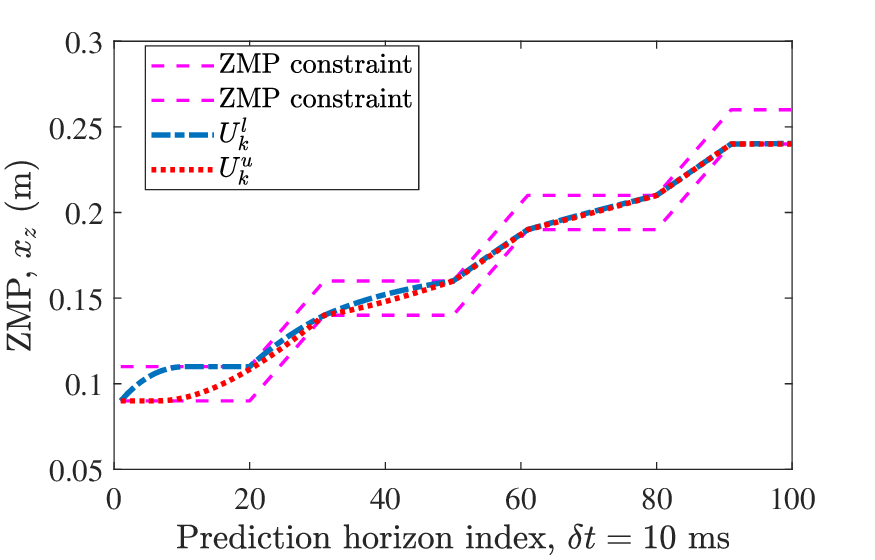}}
	\hspace{-7mm}
	\subfigure[]{
		\label{fig:ZMP:yResultRandomDisturbance} \includegraphics[width=2.43in]{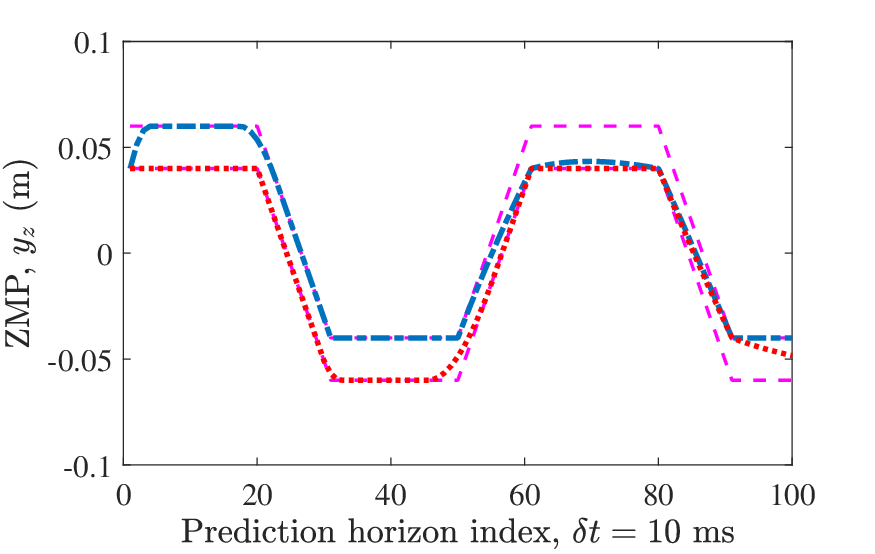}}
	\hspace{-7mm}
	\subfigure[]{
		\label{fig:ZMP:xyResultRandomDisturbance}
		\includegraphics[width=2.47in]{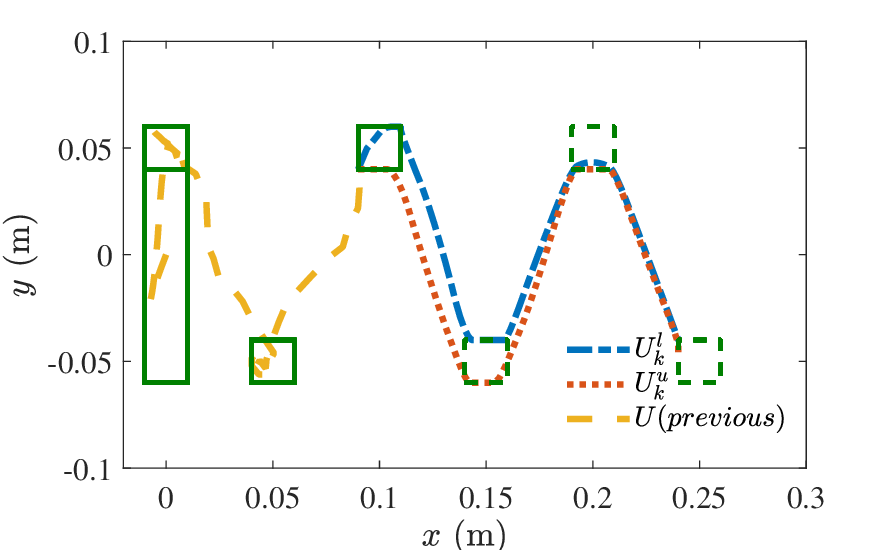}}
	\vspace{-2mm}	
	\caption{The $\mathcal{C}_1$ controls of two disturbance scenarios. ZMP trajectories w. r. t two extreme bounded disturbances, $\ddot{x}_{s}^{u}(\ddot{y}_{s}^{u})$ and $\ddot{x}_{s}^{l}(\ddot{y}_{s}^{l})$, respectively in one prediction horizon, $T_c=1$~s. Green boxes indicate ZMP geometry constraints. (a) x-t plot (b) y-t plot (c) x-y plot of ZMP under periodic sinusoidal disturbance and the prediction started at $t_k=1.02$~s. (d)-(f) Under random disturbance and the prediction started at $t_k=0.8$~s.}
\end{figure*}

\begin{figure}[h!]
\hspace{-3mm}
	\subfigure[]{
		\label{fig:COM:C1} 
		\includegraphics[width=1.7in]{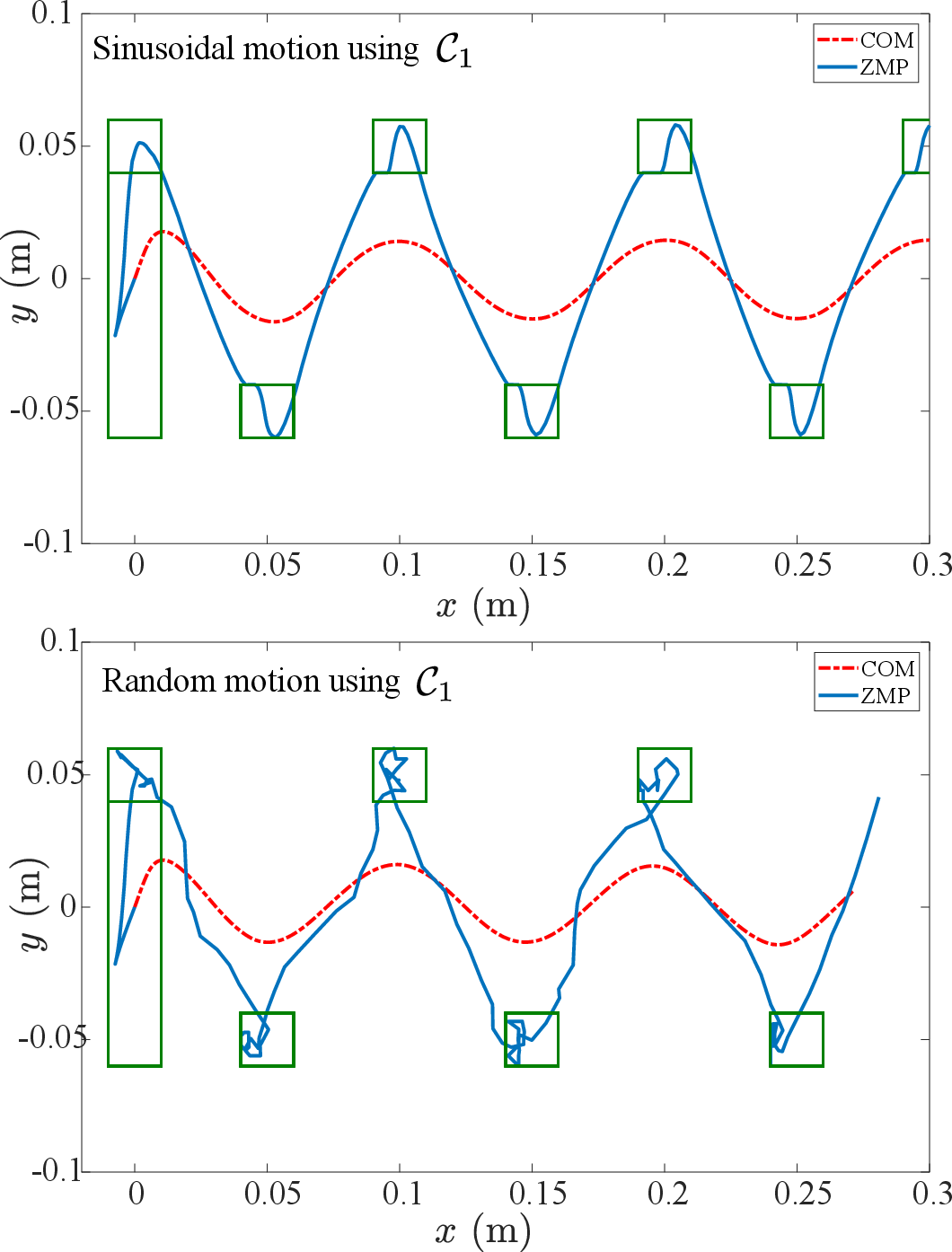}}
  \hspace{-2mm}
    \subfigure[]{
		\label{fig:COM:C2}
\includegraphics[width=1.67in]{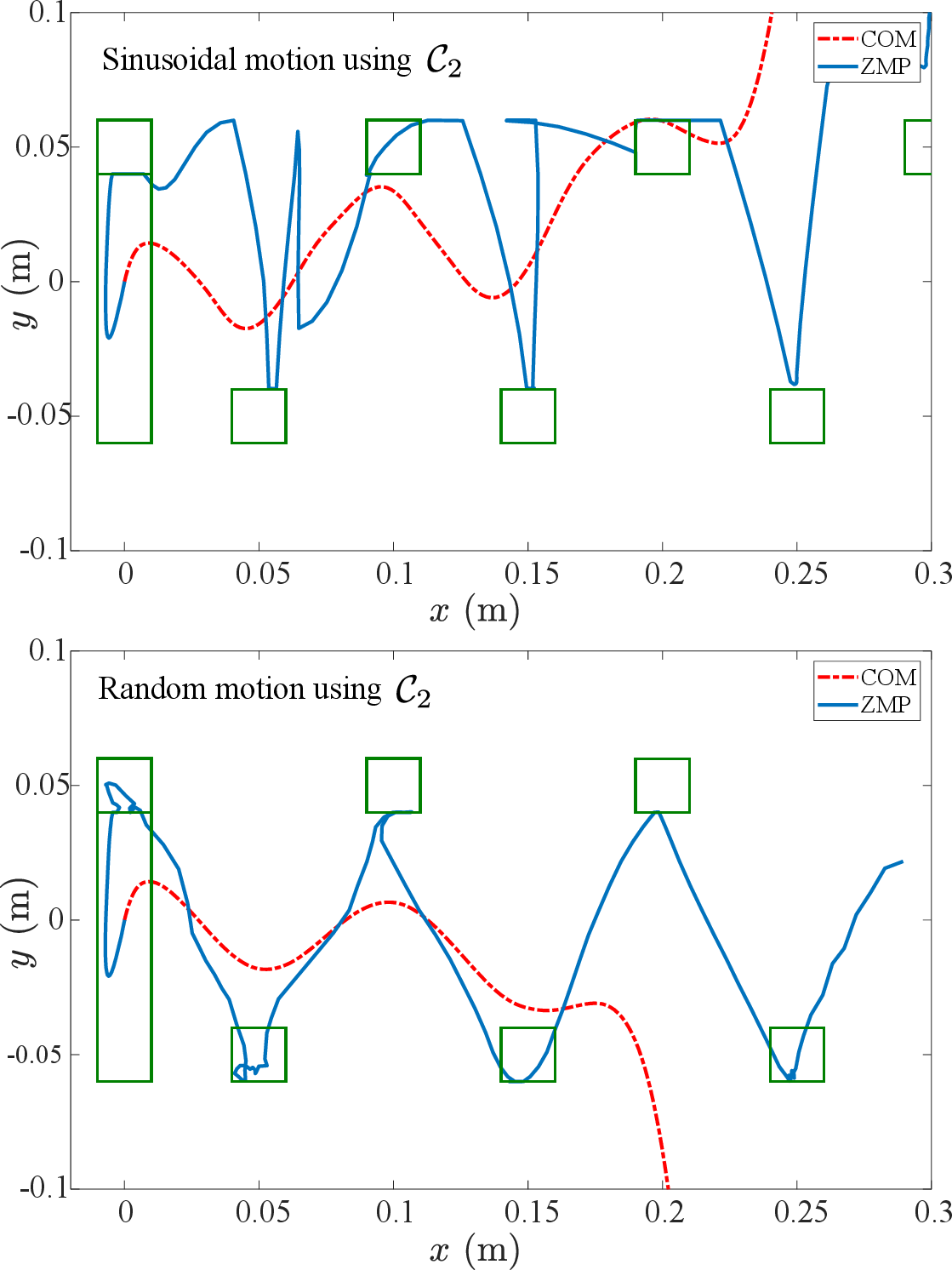}}
\vspace{-1mm}
\caption{The walking results of two scenarios. (a) Using $\mathcal{C}_1$. Top is sinusoidal motion and the bottom is random motion. (b) Using $\mathcal{C}_2$. Top is sinusoidal motion and the bottom is random motion. Blue lines are ZMP inputs and red lines are COM trajectories.}
        \label{fig:COM}
        \vspace{-3mm}
\end{figure}

Two types of the surface motion were introduced in simulation. Fig.~\ref{fig:disturbances} shows these surface motion profiles. Similar to~\cite{gao2023time}, we considered continuous and periodic acceleration disturbances. Fig.~\ref{fig:sineDisturbance} shows the disturbance profiles, that is, $\ddot{x}_s(t)=-0.04\sin[6.28(t-0.4)]$~m/s$^2$ and $\ddot{y}_s(t)=-0.22\sin{[10.47(t-0.4)]}$~m/s$^2$ in the $x$ and $y$ direction, respectively. We chose disturbance frequency $\omega_s=6.28$~rad/s ($f_s=1$~Hz) for the $x$ direction and frequency $\omega_s=10.47$~rad/s ($f_s=1.66$~Hz) for $y$ direction, respectively, as one example to differentiate two controllers. The amplitudes of these disturbances were selected to demonstrate the control performance. Since we fixed the certain foot placement to enforce the walker to step forward, it was more challenging to maintain the stability in the sagittal plane than that in frontal plane. Therefore, a smaller amplitude was used for $x_s(t)$ than $y_s(t)$. Figs.~\ref{fig:ZMP:xResultSinDisturbance}-\ref{fig:ZMP:xyResultSinDisturbance} show the predicted ZMP generation at $t_k=1.02$~s under $\mathcal{C}_1$. Fig.~\ref{fig:COM} shows the comparison between $\mathcal{C}_1$ and  $\mathcal{C}_2$. It is found that $\mathcal{C}_1$ still handled the gait under the continuous surface motion, while $\mathcal{C}_2$ failed with a divergent ZMP trajectory.

We also conducted extensive simulations to investigate the stable walking gait margin under each controller by increasing the amplitude and frequency of the sinusoidal disturbances until the robot cannot maintain the stable walking gaits. The sinusoidal motions were $x_s(t) = A_x\sin(2\pi f_{s,x} t),~y_s(t) = A_y\sin(2\pi f_{s,y} t)$. Then acceleration amplitudes of the surface in the two directions were $A_x(2\pi f_{s,x})^2$ and $A_y(2\pi f_{s,y})^2$, respectively. Table~\ref{tab:margin} lists the stable walking gait margin results. We see that under the CMPC $\mathcal{C}_1$, the robot can be stabilized successfully on moving surfaces with a comparatively larger amplitude than these under $\mathcal{C}_2$. However, the difference in disturbance frequency was not significant between $\mathcal{C}_1$ and $\mathcal{C}_2$.

\renewcommand{\arraystretch}{1.3}
\setlength{\tabcolsep}{0.1in}
\begin{table}[t!]
\vspace{1mm}
\centering
\caption{Calculated stable walking gait margin of $\mathcal{C}_1$ and $\mathcal{C}_2$ under sinusoidal disturbances.}
\label{tab:margin}
\begin{tabular}{ccccc}
\hline\hline
\multirow{2}{*}{}                       & \multicolumn{2}{l}{$x$ direction} & \multicolumn{2}{l}{$y$ direction} \\ \cline{2-5}
                                        & $\mathcal{C}_1$              & $\mathcal{C}_2$              & $\mathcal{C}_1$              & $\mathcal{C}_2$             \\ \hline
\multicolumn{1}{c}{Amplitude ($f_s=1.25$~Hz) (m/s$^2$)} & $0.15$             & $0.11$             & $0.46$             & $0.14$             \\
\multicolumn{1}{c}{Frequency ($A_{x,y}=0.5$~cm) (Hz)}      & $1.00$             & $0.80$             & $1.39$             & $1.05$             \\ \hline\hline
\end{tabular}
\vspace{-0mm}
\end{table}

We further tested the control performance when the surface moved with a continuously random acceleration. Fig.~\ref{fig:randomDisturbance} shows the disturbance profiles that were generated randomly with bounds, namely, $\ddot{x}_{s} \in [-0.5,0.5]$~m/s$^2$ and $\ddot{y}_{s} \in [-0.75,0.75]$~m/s$^2$. The surface began to shake at $t=0.3$~s. Figs.~\ref{fig:ZMP:xResultRandomDisturbance}-\ref{fig:ZMP:xyResultRandomDisturbance} show the ZMP profile from $t_k=0.8$~s. As shown in the bottom subfigure of Fig.~\ref{fig:COM:C1}, under $\mathcal{C}_1$, the bipedal walker still moved forward and was stabilized though the controlled ZMP trajectory became not smooth as those under other disturbances. The regular stable MPC $\mathcal{C}_2$ only stabilized the walker for the first two steps and then the robot fell rightward.

Within the prediction horizon $T_c$, the computed commands under CMPC, $\bs{U}^l_k$ and $\bs{U}^u_k$, were close to each other observed as in the later section of the horizon as shown in  Figs.~\ref{fig:ZMP:xResultSinDisturbance}-\ref{fig:ZMP:xyResultRandomDisturbance}. This was due to the designed CMPC stability constraint~\eqref{eqn:Stability_constraint}. To further illustrate this, the elements $e^{-\omega i \delta t}(1-e^{-\omega \delta t})$ of vector $\bs{A}_k$, $i=0,\cdots,N-1$, become close to zero as the horizon index $i$ increases, which implies that the stability constraint does not strongly influence the decision variables when index $i$ is large. But we only use the first term $x_{z,k}$ as the control input, so it is not influential.

Although the CMPC successfully demonstrated the superior performance to handle the bipedal walking on the moving surface than the regular MPC, there are some limitations for further improvement. For example, this work did not consider the vertical motion of the walking surface, nor for the COM. Furthermore, the control systems design was built on the LIP model and should be extended to integrate with whole-body control, such as those in~\cite{gao2023time,iqbal2021extended}. 

\vspace{-0mm}
\section{Conclusion and Future Work}
\label{Sec:conclusion}

In this paper, we adopted a contingency MPC concept to handle the uncertainty of the surface's motion for the bipedal walking control. A linear inverted pendulum model was built with consideration of the unknown acceleration of the moving surface. Stability analysis was conducted and a contingency MPC receding optimization problem was formulated by incorporating worst cases of the predictive surface motion. Simulation results confirmed the feasibility and superior performance of the contingency MPC design over the regular MPC. As an ongoing effort, we plan to extend the work by integrating this contingency MPC design with (1) adaptive foot placement (2) whole-body joint torque controller for bipedal walking locomotion. Experimental validation and demonstration with bipedal robots is another ongoing research direction. Further extension to exoskeleton-assisted human walking on moving surfaces (e.g.,~\cite{ZhuRAL2023}) is also among the future research direction.   

\vspace{-0mm}

\bibliographystyle{IEEEtran}
\bibliography{ChenRef}

\end{document}